\documentclass{article}

\usepackage{microtype}
\usepackage{graphicx}
\usepackage{subfigure}
\usepackage{booktabs} 

\usepackage{hyperref}

\usepackage[accepted]{icml2023}

\usepackage{amsmath}
\usepackage{amssymb}
\usepackage{mathtools}
\usepackage{amsthm}

\usepackage[capitalize,noabbrev]{cleveref}

\theoremstyle{plain}

\theoremstyle{definition}

\theoremstyle{remark}

\usepackage[textsize=tiny]{todonotes}

\icmltitlerunning{Addressing Racial Bias in Facial Emotion Recognition}

\begin{document}

\twocolumn[
\icmltitle{Addressing Racial Bias in Facial Emotion Recognition}



\icmlsetsymbol{equal}{*}

\begin{icmlauthorlist}
\icmlauthor{Alex Fan}{Stanford}
\icmlauthor{Xingshuo Xiao}{ICME}
\icmlauthor{Peter Washington}{University of Hawaii}

\end{icmlauthorlist}

\icmlaffiliation{Stanford}{Department of Statistics, Stanford University, Stanford, CA, USA}
\icmlaffiliation{ICME}{Institute of Computational Math and Engineering, Stanford University, Stanford, CA, USA}
\icmlaffiliation{University of Hawaii}{Information and Computer Science, University of Hawai'i at Manoa, Honolulu, HI, USA}

\icmlcorrespondingauthor{Alex Fan}{alexfan@stanford.edu}
\icmlcorrespondingauthor{Xingshuo Xiao}{xingshuo@stanford.edu}
\icmlcorrespondingauthor{Peter Washington}{pyw@hawaii.edu}

\icmlkeywords{Facial Emotion Recognition, AffectNet, CAFE, Racial Bias, Bias Simulation}

\vskip 0.3in
]



\printAffiliationsAndNotice{}  

\begin{abstract}
Fairness in deep learning models trained with high-dimensional inputs and subjective labels remains a complex and understudied area. Facial emotion recognition, a domain where datasets are often racially imbalanced, can lead to models that yield disparate outcomes across racial groups. This study focuses on analyzing racial bias by sub-sampling training sets with varied racial distributions and assessing test performance across these simulations. Our findings indicate that smaller datasets with posed faces improve on both fairness and performance metrics as the simulations approach racial balance. Notably, the F1-score increases by $27.2\%$ points, and demographic parity increases by $15.7\%$ points on average across the simulations.  
However, in larger datasets with greater facial variation, fairness metrics generally remain constant, suggesting that racial balance by itself is insufficient to achieve parity in test performance across different racial groups.
\end{abstract}

\section{Introduction}
\label{submission}
Emotion recognition, commonly referred to as facial expression recognition (FER), encompasses the identification and analysis of facial expressions displayed by individuals in images or videos. 
This complex procedure consists of three key stages: face detection, feature extraction, and emotion classification. \cite{Ko2018ABR}. 
FER finds extensive utility across various domains including human-computer interaction (HCI) \cite{picard1999affective}, media analytics \cite{zhao2019affective}, robotics \cite{tao2005affective}, and health informatics \cite{voss2019effect, washington2022improved}.

Historically, automatic emotion recognition predominantly relied on the extraction of domain-specific features such as facial action units \cite{AU}. However, with the rapid progression of machine learning (ML) and deep learning (DL) techniques, deep neural networks (DNNs) have emerged as a prominent approach for developing facial emotion recognition models. Such models necessitate expansive datasets to ensure robustness and accuracy in their predictions.
In recent years, researchers have proposed DL models which leverage more expansive datasets for training \cite{FERsurvey}. 
Despite the impressive achievements of deep learning methods in FER, a significant challenge arises from the presence of racial bias in such models. This issue, which is well documented in existing literature \cite{chen2021understanding, domnich2021responsible, sham2023ethical, xu2020investigating}, necessitates immediate attention to mitigate discriminatory outcomes and to provide equitable opportunities for individuals of diverse ethnicities and skin colors. Addressing and rectifying the biases within DNNs is of paramount importance for real-world translation of such models.

Biases within DNNs can primarily be attributed to two fundamental sources: the training data and the algorithms themselves \cite{biasDLSurvey}. Given that models learn from input data, any biases present within the underlying datasets are inherently ingrained within the learning process of the algorithms. 
Furthermore, the design of feature extraction processes for these models may introduce biases that disproportionately affect different racial groups. An illustrative example is the consideration of skin color as a learned feature extracted during the deep learning process, which can ultimately lead to unfair predictions.


To address the issue of racial bias in FER datasets, we conduct a simulation study on AffectNet and CAFE datasets \cite{affectnet_data, CAFE_data}.
In each simulation, we select a specific race as the simulated race, and ensure other races equal representations. We train the FER model using sub-sampled data with varying proportions of the simulated race, measuring the accuracy, F1-scores, and fairness metrics. We find that the racial balance of the training set has some influence on the test race-specific F1-score, but mitigating balance alone is insufficient to address other types of bias such as annotator bias.

\section{Related Works}
\subsection{Deep-learning-based Emotion Recognition}
Deep learning models serve as the foundation for each stage of FER \cite{Ko2018ABR}. 
Training DNNs for FER requires the utilization of diverse datasets, encompassing varying numbers of labels and data types \cite{affectnet_data, CAFE_data, CK_data, JAFFE_data, RAF-DB, FER2013}. 
Li and Deng provided an overview of popular datasets designed specifically for deep emotion recognition \yrcite{FERsurvey}. 

In the realm of deep-learning-based FER approaches, convolutional neural networks (CNNs) and recurrent neural networks (RNNs) are frequently employed. 
Khorrami et al. demonstrated that CNNs excel at accurately extracting facial action units (FAUs), thereby yielding promising classification performance on the CK+ dataset \yrcite{khorrami2015deep}. 
Lopes et al. proposed a combination of CNNs and image pre-processing techniques, such as cropping and normalization, to reduce the number of convolutional layers and alleviate the need for extensive training data. This approach resulted in improved overall accuracy and computational efficiency on the CK+, JAFFE, and BU3DFE datasets \yrcite{LOPES2017610}. 
Agrawal and Mittal introduced a novel CNN model that investigated the influence of kernel size and the number of filters on the final classification accuracy, leading to further improvements in performance 
 \yrcite{agrawal2020}. 
Recent studies have also explored the integration of generative adversarial networks (GANs) within CNNs for data augmentation and training purposes \cite{peng2020, porcu}.

Incorporating CNNs with RNNs, Zhu et al. proposed a novel FER approach that integrates features learned from each layer of a CNN within a bidirectional RNN architecture \yrcite{Zhu2017}. 
RNNs have also proven effective in capturing dynamic facial actions within multimodal FER, as demonstrated by Majumder et al. \cite{majumder2019dialoguernn, Hasani_2017}.

Despite achieving high overall accuracy, the performance across different racial groups has yet to be thoroughly and systematically studied. While multicultural FER models have demonstrated their effectiveness in improving accuracy, their impact on fairness has yet to be explicitly addressed \cite{sohail, ali2016boosted}.

\subsection{Fairness in Machine Learning}
Numerous research endeavors have been dedicated to addressing issues of unfairness within ML systems \cite{fairDL, biasDLSurvey, Oneto_Chiappa_2020, Mehrabi_2022}.
The underlying causes of biases in ML have been explored in several publications \cite{chouldechova2018frontiers, Martínez_2019}. 
Evaluation metrics, shaped by social contexts, are employed to gauge the fairness of these models \cite{chouldechova2018frontiers, castelnovo2021zoo}.

ML fairness improvement methods are generally categorized into pre-processing, in-processing, and post-processing techniques, contingent upon the stage at which the fairness correction method is applied  \cite{fairDL, biasDLSurvey, Oneto_Chiappa_2020, Mehrabi_2022}.
Additionally, AI fairness toolkits have been developed, harnessing these methods \cite{bird2020fairlearn, AIF360, what_if, 2018aequitas}.  

Several prior studies have delved into the examination and alleviation of racial bias in FER. 
Raina et al. utilized artificial facial images and observed racial bias in FER models \yrcite{raina2022exploring}. Sham et al. conducted an investigation into racial bias in popular state-of-the-art FER methods and revealed that the presence of uneven or insufficient representation within the training data leads to biased performance outcomes \yrcite{FERbiasrace}.

Additionally, biases in expression labeling within datasets, influenced by the impact of races on emotion perceptions, was identified as contributors to unfairness \cite{Rhue_2018, annotateBias}. 
Conversely, Chen and Joo's study did not report any systematic labeling biases for races, attributing the absence to imbalanced racial representations in the dataset \yrcite{annotateBias}. 

Although some methods have demonstrated effectiveness in correcting FER racial bias \cite{biasesStudy}, the results highly depend on the datasets and models.
Due to the involvement of highly subjective labels and high-dimensional inputs in emotion recognition, the field has yet to address the issue of fairness comprehensively in this domain and similar areas with complex and heterogeneous data streams.


\section{Methods}

\subsection{Datasets}
We employ two datasets to investigate racial bias. The first dataset is the Child Affective Facial Expression (CAFE) dataset, a collection of images featuring children posing specific emotions \cite{CAFE_data}. The second dataset is AffectNet, a widely recognized large-scale dataset for general facial emotion recognition \cite{affectnet_data}. 

To align the datasets, we filter the examples within AffectNet to include only those with emotion labels matching those in the CAFE dataset, specifically neutral, sadness, happiness, surprise, anger, disgust, and fear. Additionally, we exclude grayscale images, which also contributes to more accurate race estimates. We calculate the per-pixel squared error (summed across the three channels) using the mean pixel value of each image. Images with an average per-pixel squared error below a threshold are considered grayscale and are removed from the training set. Consequently, the final training size for AffectNet is N = 259,280. A similar procedure was followed for the validation and test sets, resulting in sizes of N = 1,700 and 1,484, respectively.

To separate the faces in CAFE from their white background, we utilize OpenCV bounding boxes, following a methodology that AffectNet uses during its data collection process \cite{affectnet_data, opencv_library}. We exclude images in which a face could not be adequately bounded, resulting in 1,178 usable images. Because the CAFE dataset involves participants posing for multiple emotions, we opt to split the data at the participant level when generating the training, validation, and test sets, resulting in sizes of N = 713, 227, and 222, respectively.

\subsection{Race Estimates}

\begin{table}[t]
\begin{center}
\caption{The racial distribution in CAFE's training set. Participants self-report their race prior to data collection. A plurality of participants identify as European-American, indicating the potential for downstream racial bias in the model. \label{table:cafe_racial_distribution}}

\vskip 0.15in
\begin{small}
\begin{sc}
\begin{tabular}{ccc}
\toprule
Race & Count & Proportion \\
\midrule

European-American & 281 & 0.404 \\
African-American& 142 & 0.205 \\
East Asian& 103 & 0.148\\
Latino & 86 & 0.124\\
South Asian& 82 & 0.118 \\
\bottomrule
\end{tabular}
\end{sc}
\end{small}
\end{center}
\vskip -0.1in
\end{table}

\begin{table}[t]
\begin{center}
\caption{The racial distribution in AffectNet's training set. AffectNet does not provide race information, so this distribution is estimated using a race model built from FairFace. European-American faces account for a much larger proportion of the dataset relative to CAFE's composition. \label{table:aff_racial_distribution}
}
\vskip 0.15in
\begin{small}
\begin{sc}
\begin{tabular}{ccc}
\toprule
Race & Count & Proportion \\
\midrule
 European-American & 174,382& 0.673 \\
African-American & 19,131& 0.074\\
East Asian & 15,833& 0.061 \\
 Latino & 23,488 &  0.091 \\
Middle Eastern & 18,120& 0.070\\
  South Asian & 4,786 & 0.018 \\
Southeast Asian& 3,540& 0.014\\

\bottomrule
\end{tabular}
\end{sc}
\end{small}
\end{center}
\vskip -0.1in
\end{table}

We require race labels to analyze the bias. CAFE provides participants' self-reported ground truth race labels: European-American, African-American, (East) Asian, Latino, and South Asian. However, for the AffectNet dataset, race information is not available; we approach this problem by estimating race labels. We utilize models trained on labeled race datasets, specifically FairFace, which exhibits greater racial balance compared to similar datasets, and models trained on FairFace demonstrate improved performance on non-white faces relative to other models and datasets \cite{karkkainenfairface}. Our model uses the paper's original weights to predict the race labels, and the counts and proportions of the two datasets are presented in Tables \ref{table:cafe_racial_distribution} and \ref{table:aff_racial_distribution}. As expected, European-American faces make up the majority of the training set distributions for both CAFE and AffectNet, comprising 40.4\% and 67.3\% of their respective datasets.

The FairFace-based model categorizes the AffectNet faces into seven categories: European-American, African-American, (East) Asian, Latino, South Asian, Middle Eastern, and Southeast Asian. It is possible to exclude the Middle Eastern and Southeast Asian groups, from the AffectNet experiments to align the racial categories with those of CAFE. However, we choose to retain all races in the experiments assuming that there could be latent information from these additional race categories that affects the training process. 

\begin{figure*}[tbp]
\centering
\includegraphics[width=.95\linewidth]{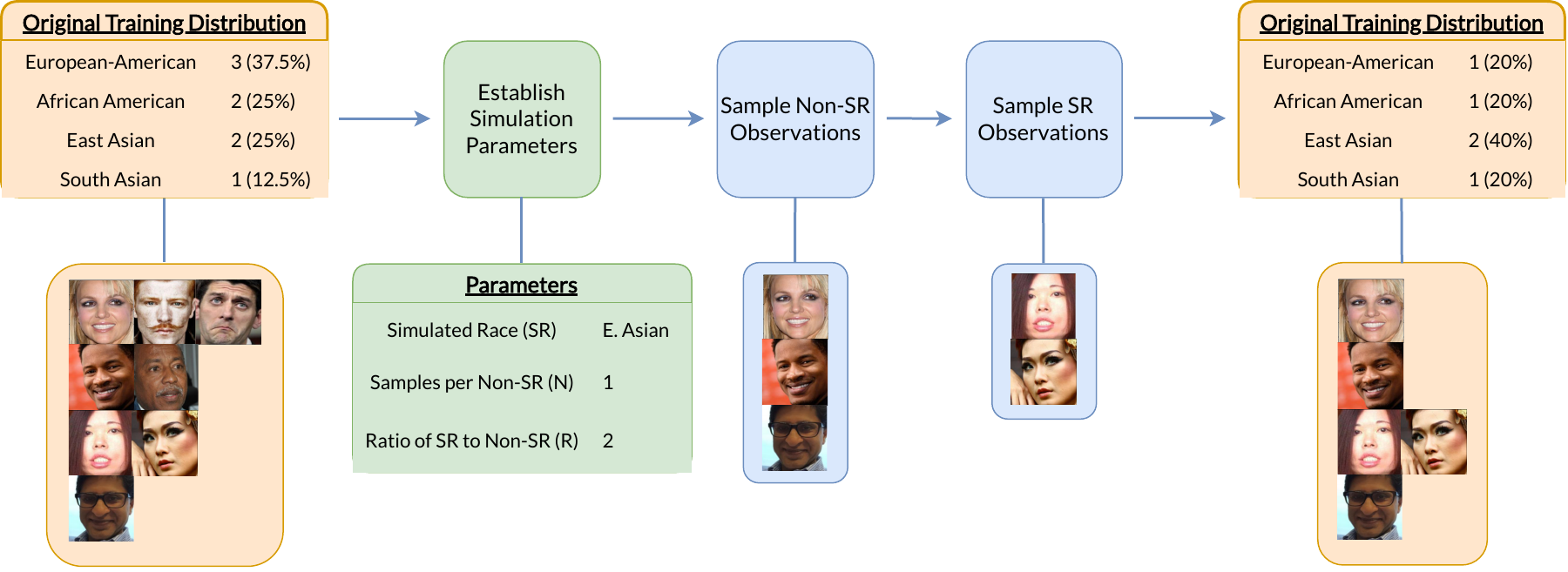}
\caption{ \label{fig:simulation_diagram}
Example of the procedure for simulating East Asian representation in a dataset. This represents a contrived case where East Asian is over-represented in the training set. In actual simulations, $N$ is much larger, with our largest experiment using $N = 3500$.}
\end{figure*}

\subsection{Simulating Racial Composition}
To investigate the impact of racial representation in the training set on test performance, we select a specific race (henceforth referred to as the simulated race) and vary their proportion. The sampling process ensures an equal representation for non-simulated races by sampling $N$ examples for each race. Then we set a ratio ($R$) of the simulated race to a non-simulated race and sample $N*R$ examples of the simulated race. All sampling is done without replacement. We analyze the effects of under-representation and over-representation of the simulated race by varying the ratio.

In the simulations, we fine-tune a ResNet-50 model on the sub-sampled training set. The model's performance is evaluated on the validation set throughout the epochs, and the weights yielding the highest accuracy on the validation set are used to evaluate on the test set. While accuracy serves as one evaluation metric, we also consider the race-specific F1-score, which is calculated after filtering to the simulated race. Additionally, we explore fairness metrics and extend their applicability to multi-class classification problems.

Demographic parity and equality of odds are two fairness principles that are commonly used in the algorithmic fairness literature \cite{barocas-hardt-narayanan}. Demographic parity ensures that the positive prediction rate remains consistent across sensitive attributes. One approach to quantify this principle is the use of the ratio of the smallest positive prediction rate to the largest. A value of one suggests that the model achieves demographic parity \cite{bird2020fairlearn}. To implement this metric, we transform the multi-class problem into multiple one-versus-rest sub-problems. For each emotion, we compute the demographic parity ratio across the different races and then average these ratios to obtain an overall measure of demographic parity.

Equality of odds requires parity in both true-positive and false-positive rates across sensitive attributes. A similar procedure to that used for the demographic parity ratio can be employed to derive two separate ratios: the true-positive parity ratio and the false-positive parity ratio \cite{bird2020fairlearn}. The equalized odds ratio is determined by selecting the smaller of the two ratios. This indicates that equality of odds is achieved when both parity ratios closely approximate one.

\section{Experiments}

We conduct three sets of simulations in our study. The first simulation focuses on the CAFE dataset, which we sample at the participant level. Given the limited size of the training set, we set the number of participants $N$ to be $5$ and the ratio $R$ ranges from 0 to 2.0 with increments of 0.2. This approach ensures that each simulation comprises a whole number of participants, and the sizes of each non-simulated races are approximately equal, with 40-50 observations per racial group. 

The second simulation involves the AffectNet dataset with $N$ set to 50 observations, and $R$ follows the same range and increments as in the previous simulation. This simulation tests the consistency of results between AffectNet and CAFE in the context of a small dataset size regime.

In the third simulation, we use the AffectNet dataset with $N$ set to 3500 observations, allowing for a larger training set. This simulation investigates whether the trends observed in the first and second simulations generalize to larger data regimes, which is relevant for the typical application of facial emotion recognition.

Throughout the experiments, we maintain consistent training hyperparameters. We use a fixed learning rate of $1e-4$ with an Adam optimizer (with $\beta_1 = 0.9$ and $\beta_2 = 0.999$) and L2-weight decay on the model parameters. Cross-entropy loss is backpropagated at each batch step, and each simulation undergoes training for 5 epochs. To account for the emotion label imbalance within the sub-sampled dataset, we apply a weight to the loss function based on the number of ground truth labels. These experiments were conducted using an NVIDIA K80 GPU unit.

\subsection{Results}

\paragraph{CAFE Simulation}

We display metrics from various levels of $R$ in four race simulations: African-American, East Asian, European-American, and Latino. We calculate weighted F1-score and accuracy at a race level by filtering the predictions to the simulated race. We calculate demographic parity ratio and equalized odds ratio without any filters. Our hypothesis anticipates that race-specific F1-score and fairness metrics would improve as the dataset becomes more racially balanced. However, as the simulations over-sample and the simulated race becomes over-represented, we expect fairness metrics to plateau or decline, as the model's fairness performance for the non-simulated races is likely to deteriorate.

\begin{figure}[btp]
\vskip 0.2in
\begin{center}
\includegraphics[width=\linewidth]{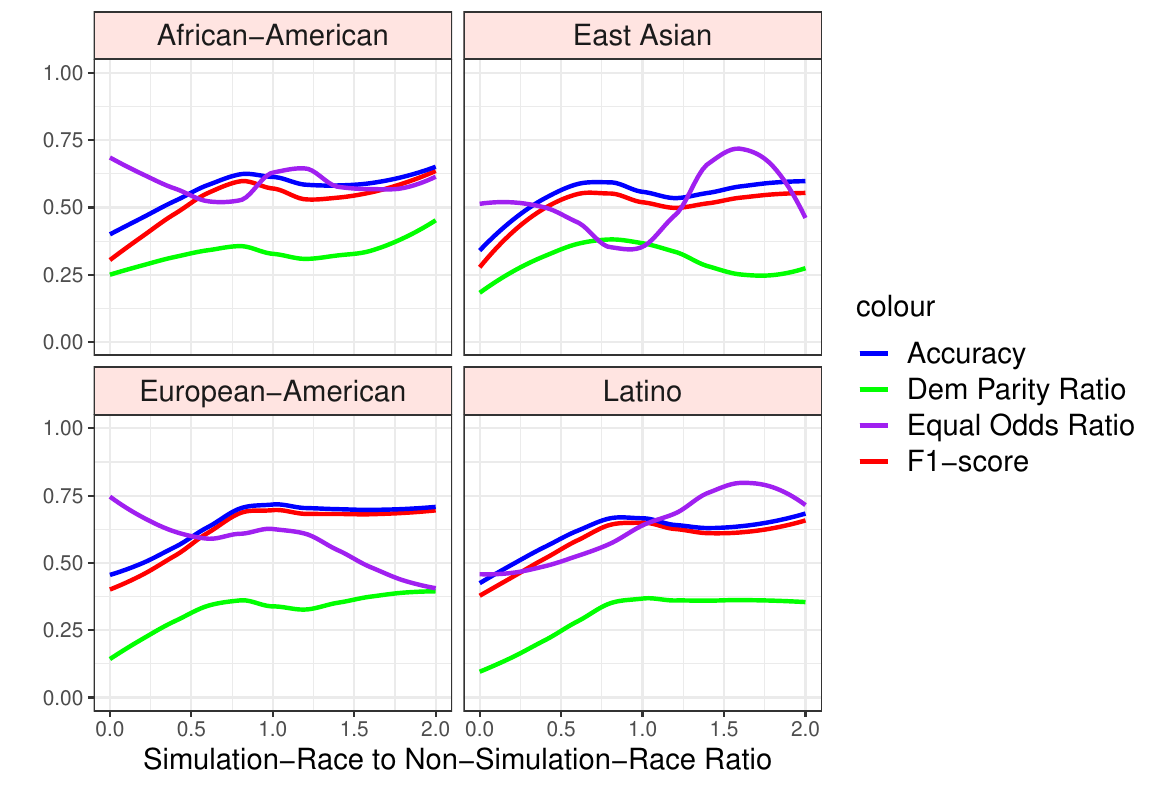}
\caption{ \label{fig:cafe_all_metrics}
CAFE racial composition simulations with all test metrics. Each cell shows a varied simulated race with all non-simulated races held constant. CAFE is sampled at the participant level ($N = 5$). Every race shows improvement in test performance for the race-specific F1-score and demographic parity ratio when the simulations move towards racial balance.}
\end{center}
\vskip -0.2in
\end{figure}
\begin{figure}[btp]
\vskip 0.2in
\begin{center}
\includegraphics[width=\linewidth]{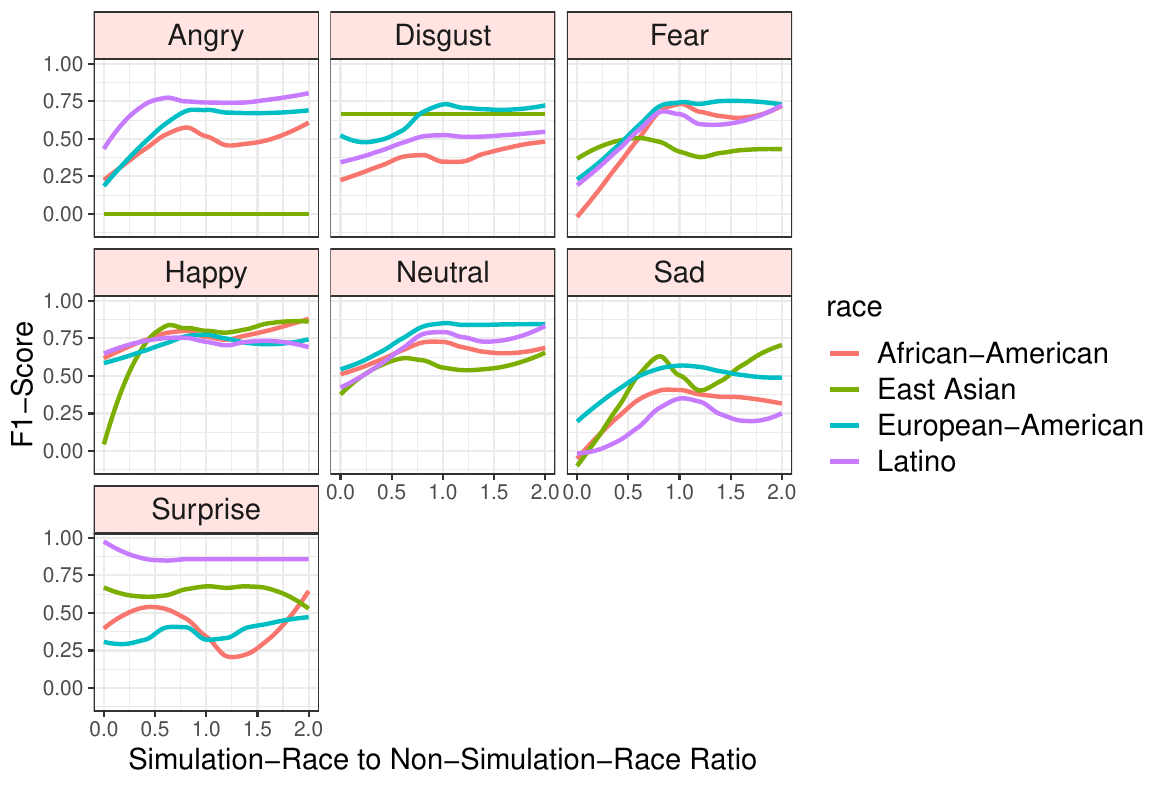}
\caption{ \label{fig:cafe_f1_score_unagg}
CAFE racial composition simulations with unaggregated F1-scores show racial balance is correlated with increases in most emotion-specific F1-scores. The exceptions are `surprise' and `disgust'.}
\end{center}
\vskip -0.2in
\end{figure}

The simulations conducted on the CAFE dataset align with expectations on some metrics. Figure \ref{fig:cafe_all_metrics} shows that the F1-score and the demographic parity ratio increase ($+27.2\%$ and $+15.7\%$ points respectively on average) as the dataset becomes balanced and stabilize when the dataset over-samples the simulated race. On the other hand, the equalized odds ratio exhibits greater inconsistency, with only the Latino simulations displaying a clear upward trend, while the other races exhibit random or downward trends.

Additionally, in Figure \ref{fig:cafe_f1_score_unagg} we present the disaggregated F1-scores for each race and label before their aggregation. The disaggregated results reveal that a significant portion of the F1-score improvement stems from emotions such as neutral, sad, and fear. Interpreting changes in East Asian F1-scores proves challenging, possibly due to the limited presence of Asian participants in the test set. Moreover, surprise and disgust appear to be more challenging emotions to predict, which could explain the seemingly random or marginal trends observed.

\paragraph{AffectNet Small Simulation}

\begin{figure}[btp]
\vskip 0.2in
\begin{center}
\includegraphics[width=1\linewidth]{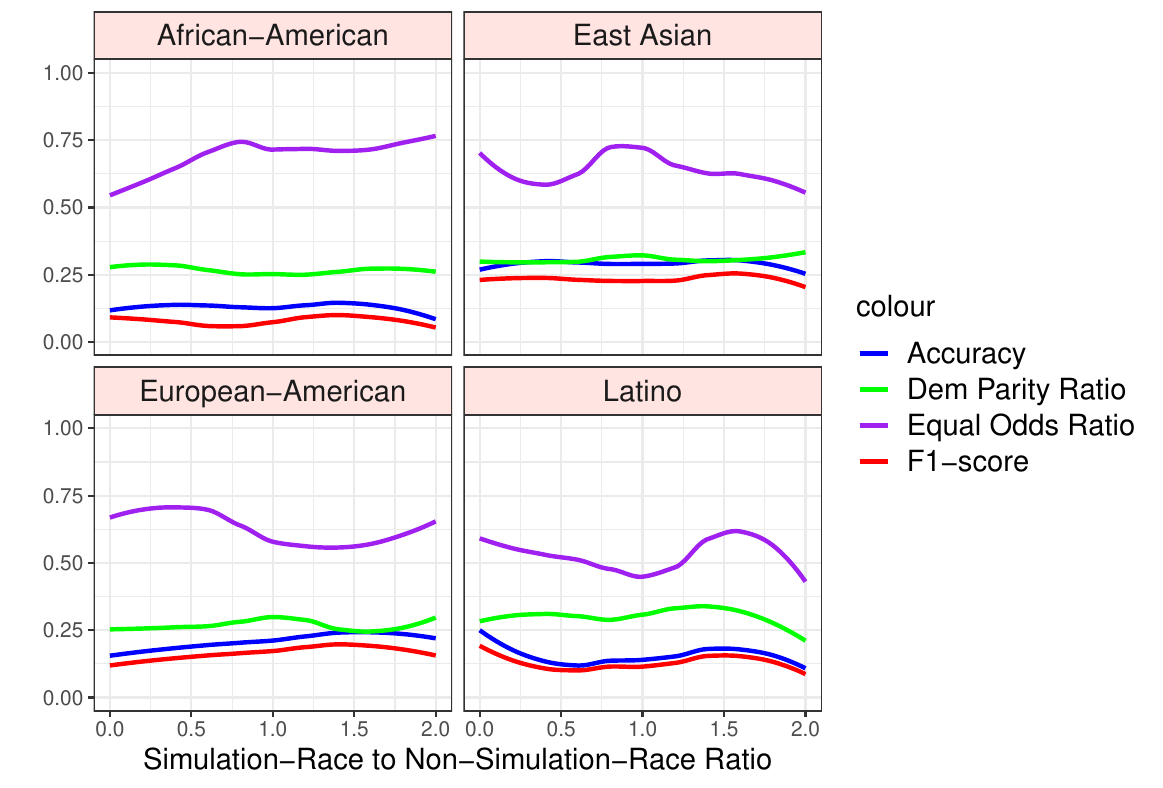}
\caption{ \label{fig:aff_small_all_metrics}
AffectNet racial composition simulations using small subsample sizes ($N = 50$) with all test metrics. Race-specific F1-score and demographic parity ratio have poor performance likely due to overfitting, and the lack of an observable trend in any of the simulations suggests minimal correlation between racial balance and test performance.}
\end{center}
\vskip -0.2in
\end{figure}

The performance on the small sub-sample of AffectNet, achieving 15.2\% race-specific F1-score and 0.286 demographic parity ratio on average, is noticeably inferior to the CAFE simulations. The limited size of the training set and the substantially greater variation in emotion distribution from the ``wild'' images in AffectNet likely contribute to this discrepancy. Despite the potential for model overfitting, the overall trend shown in Figure \ref{fig:aff_small_all_metrics} indicates that the model's performance does not significantly change as the dataset becomes more racially balanced, as evidenced by the nearly random trends observed in the F1-score and fairness metrics.

\paragraph{AffectNet Large Simulation}

\begin{figure}[t]
\vskip 0.2in
\begin{center}
\includegraphics[width=1\linewidth]{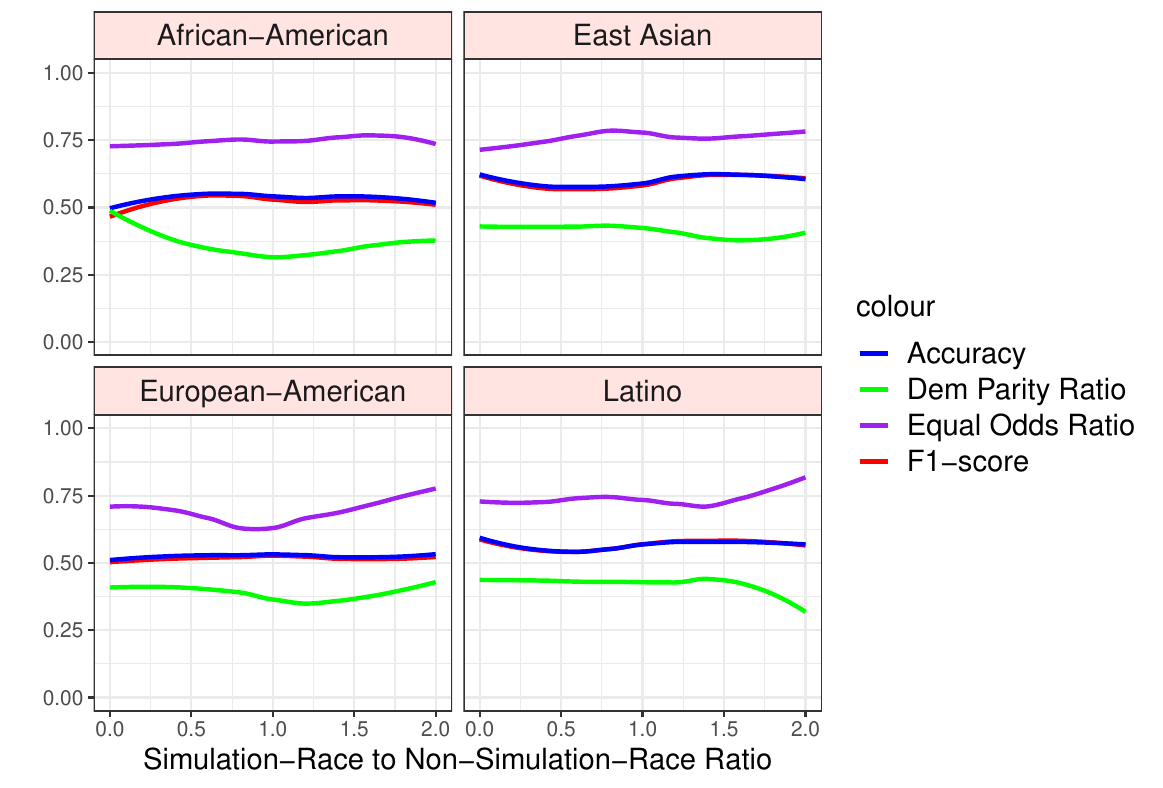}
\caption{ \label{fig:aff_larger_all_metrics}
AffectNet racial composition simulations using larger subsample sizes ($N = 3500$) with all test metrics. Although the larger training set resolves overfitting to a degree, the simulations still lack a visible trend between racial balance and test performance.}
\end{center}
\vskip -0.2in
\end{figure}

The performance on the larger AffectNet simulation, although overfitting less, does not exhibit increases in F1-score and fairness metrics as the dataset becomes racially balanced. Figure \ref{fig:aff_larger_all_metrics} shows the race-specific F1-score fluctuating around 55\% on average for all race simulations with no visible trends. Both demographic parity ratio and equalized odds ratio also stay roughly constant or even trend downwards. 

Furthermore, the unaggregated F1-scores demonstrate minimal variation, with the exception of anger, which experiences a temporary improvement when the Asian simulations are over-sampled.

\begin{figure}[t]
\vskip 0.2in
\begin{center}
\includegraphics[width=1\linewidth]{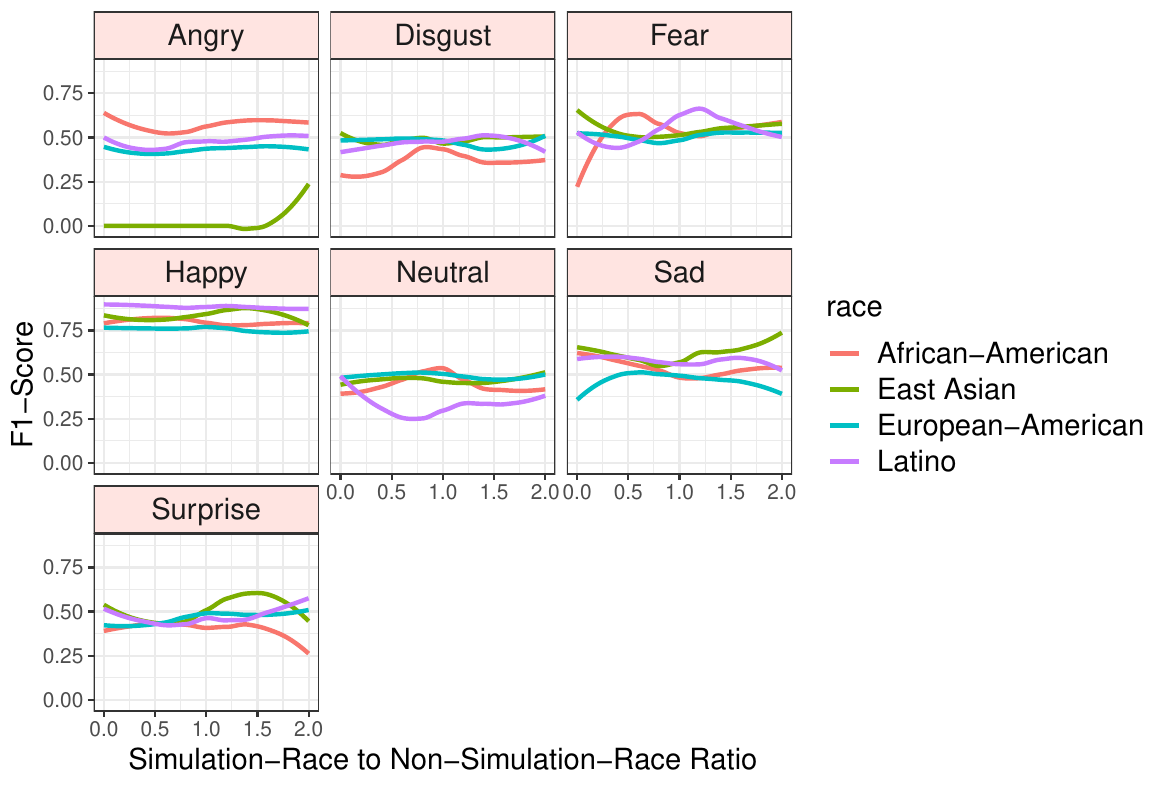}
\caption{ \label{fig:aff_larger_f1_score_unaggregated}
The unaggregated F1-scores for the AffectNet simulations show similar constant trends regardless of racial composition. `Angry' has a moderate increase in the East Asian simulations, but only when they are over-sampled.}
\end{center}
\vskip -0.2in
\end{figure}

\begin{figure}[tbp]
\begin{center}
\underline{\textbf{European-American}} \\
\includegraphics[width=1\linewidth]{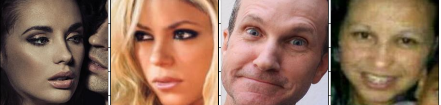} \\
\underline{\textbf{African-American}}\\
\includegraphics[width=1\linewidth]{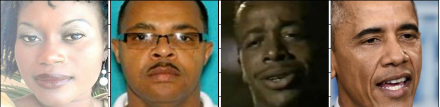}\\
\underline{\textbf{East Asian}}\\
\includegraphics[width=1\linewidth]{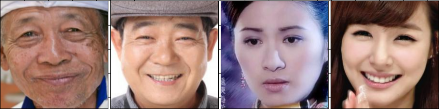}\\
\underline{\textbf{Latino}}\\
\includegraphics[width=1\linewidth]{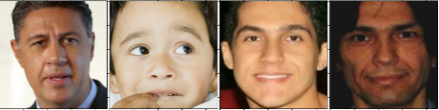}\\
\underline{\textbf{South Asian}}\\
\includegraphics[width=1\linewidth]{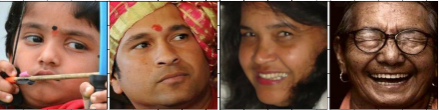}\\
\underline{\textbf{Middle Eastern}}\\
\includegraphics[width=1\linewidth]{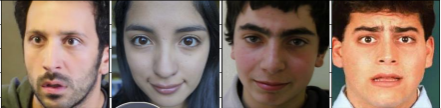}\\
\underline{\textbf{Southeast Asian}}\\
\includegraphics[width=1\linewidth]{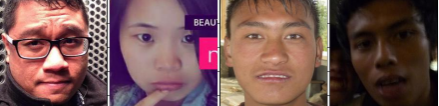}\\
\caption{ \label{fig:race_batch}
A random sample of image/race examples from the FairFace race model estimates. Qualitatively, estimates of European-American, African-American, East Asian, and Latino individuals appear more accurate than of South Asian, Middle Eastern and Southeast Asian individuals. }
\end{center}
\end{figure}
\section{Discussion}
The simulations conducted on CAFE indicate that racial balance of the training set has some influence on the test race-specific F1-scores. However, only certain fairness metric trends align with our expectations. Demographic parity ratio increases as the training set becomes balanced and then plateaus, but equalized odds ratio, a stricter metric, shows random trends through different racial distribution simulations.

The AffectNet simulations exhibit mostly random trends across scenarios. Nevertheless, even for CAFE, biases persist within the models since the fairness metrics fail to approach a value of 1. This suggests that there may be sources of bias that the simulations are unable to capture. One possible explanation could be biased race estimations obtained from the FairFace model. Upon examining a small sample of faces categorized by this model, we observe that White, Black, Latino, and East Asian faces are estimated accurately. Other race and ethnicity categories, however, appear to be more prone to misclassification, particularly Middle Eastern faces based on lighter skin tones and Indian faces based on darker skin tones. This bias could potentially impact the training process even though the simulations focus on races estimated with reasonable accuracy. However, since the race labels for AffectNet are not available, it is challenging to ascertain the extent of estimator bias. As part of future extensions, it may be worthwhile to exclude the less-consistent racial categories to explore whether the estimator bias contributes to the lack of trends in the AffectNet simulation.

Another source of bias that cannot be addressed through the simulation of racial compositions is annotation bias. There is evidence in the psychology literature suggesting that individuals are less accurate in determining facial expressions for races different from their own, resulting in potentially disproportionate labeling biases \cite{zhongqing_jiang_other-race_2023, annotateBias}. Annotation bias could be particularly problematic for AffectNet since its annotations were derived from only 12 labelers, with most labels being annotated by a single individual\footnote{The authors note that a two annotators labeled a subset of the data and there is agreement between the two annotators ranging from 50.8\% on neutral labels to 79.6\% on happy labels.} \cite{affectnet_data}. It would be possible to quantify and analyze this bias by collecting data on each observation, the race of the labelers, and the annotation process. CAFE, in fact, provides information about the aggregate race distribution of the labelers \cite{CAFE_data}, and this information has been incorporated into the training process in prior work \cite{washington2021training}. 

Given the persistence of these non-compositional biases within AffectNet, traditional bias mitigation techniques like loss re-weighting or fairness regularization may not yield strong results. This underscores the need for the fairness community to explore alternative methods of bias mitigation, particularly in settings involving high-dimensional image inputs and subjective labels.

\section{Conclusion}
There is an ongoing need for addressing fairness in facial emotion recognition. By simulating different racial distributions within the training sets, we demonstrate the impact of compositional racial imbalance on test performance and disparities in the CAFE dataset. Moreover, we extend this analysis to AffectNet, which comprises non-posed, ``in the wild'' expressions. The results reveal the persistent presence of bias across simulations of racial composition in the training set, with no improvement observed in the performance of race-specific F1-scores even when enforcing racial balance within the simulation. To further advance research in this domain, we propose exploring additional avenues of inquiry, such as re-simulating the AffectNet experiments while excluding racial groups that are inaccurately estimated during the pre-processing stage.


\bibliography{icml2023}

\begin{thebibliography}{49}
\providecommand{\natexlab}[1]{#1}
\providecommand{\url}[1]{\texttt{#1}}
\expandafter\ifx\csname urlstyle\endcsname\relax
  \providecommand{\doi}[1]{doi: #1}\else
  \providecommand{\doi}{doi: \begingroup \urlstyle{rm}\Url}\fi

\bibitem[Agrawal \& Mittal(2020)Agrawal and Mittal]{agrawal2020}
Agrawal, A. and Mittal, N.
\newblock Using cnn for facial expression recognition: a study of the effects
  of kernel size and number of filters on accuracy.
\newblock \emph{The Visual Computer}, 36\penalty0 (2):\penalty0 405--412, 2020.

\bibitem[Ali et~al.(2016)Ali, Iqbal, and Choi]{ali2016boosted}
Ali, G., Iqbal, M.~A., and Choi, T.-S.
\newblock Boosted nne collections for multicultural facial expression
  recognition.
\newblock \emph{Pattern Recognition}, 55:\penalty0 14--27, 2016.

\bibitem[Barocas et~al.(2019)Barocas, Hardt, and
  Narayanan]{barocas-hardt-narayanan}
Barocas, S., Hardt, M., and Narayanan, A.
\newblock \emph{Fairness and Machine Learning: Limitations and Opportunities}.
\newblock fairmlbook.org, 2019.
\newblock \url{http://www.fairmlbook.org}.

\bibitem[Bellamy et~al.(2018)Bellamy, Dey, Hind, Hoffman, and et~al.]{AIF360}
Bellamy, R. K.~E., Dey, K., Hind, M., Hoffman, S.~C., and et~al., S.~H.
\newblock {AI} fairness 360: An extensible toolkit for detecting,
  understanding, and mitigating unwanted algorithmic bias.
\newblock \emph{CoRR}, abs/1810.01943, 2018.
\newblock URL \url{http://arxiv.org/abs/1810.01943}.

\bibitem[Bird et~al.(2020)Bird, Dud{\'i}k, Edgar, Horn, Lutz, Milan, Sameki,
  Wallach, and Walker]{bird2020fairlearn}
Bird, S., Dud{\'i}k, M., Edgar, R., Horn, B., Lutz, R., Milan, V., Sameki, M.,
  Wallach, H., and Walker, K.
\newblock Fairlearn: A toolkit for assessing and improving fairness in {AI}.
\newblock Technical Report MSR-TR-2020-32, Microsoft, May 2020.

\bibitem[Bradski(2000)]{opencv_library}
Bradski, G.
\newblock {The OpenCV Library}.
\newblock \emph{Dr. Dobb's Journal of Software Tools}, 2000.

\bibitem[Castelnovo et~al.(2021)Castelnovo, Crupi, Greco, Regoli, Penco, and
  Cosentini]{castelnovo2021zoo}
Castelnovo, A., Crupi, R., Greco, G., Regoli, D., Penco, I.~G., and Cosentini,
  A.~C.
\newblock The zoo of fairness metrics in machine learning.
\newblock 2021.

\bibitem[Chen \& Joo(2021{\natexlab{a}})Chen and Joo]{annotateBias}
Chen, Y. and Joo, J.
\newblock Understanding and mitigating annotation bias in facial expression
  recognition.
\newblock \penalty0 (arXiv:2108.08504), Aug 2021{\natexlab{a}}.
\newblock URL \url{http://arxiv.org/abs/2108.08504}.

\bibitem[Chen \& Joo(2021{\natexlab{b}})Chen and Joo]{chen2021understanding}
Chen, Y. and Joo, J.
\newblock Understanding and mitigating annotation bias in facial expression
  recognition.
\newblock In \emph{Proceedings of the IEEE/CVF International Conference on
  Computer Vision}, pp.\  14980--14991, 2021{\natexlab{b}}.

\bibitem[Chouldechova \& Roth(2018)Chouldechova and
  Roth]{chouldechova2018frontiers}
Chouldechova, A. and Roth, A.
\newblock The frontiers of fairness in machine learning.
\newblock \emph{arXiv preprint arXiv:1810.08810}, 2018.

\bibitem[Domnich \& Anbarjafari(2021)Domnich and
  Anbarjafari]{domnich2021responsible}
Domnich, A. and Anbarjafari, G.
\newblock Responsible ai: Gender bias assessment in emotion recognition.
\newblock \emph{arXiv preprint arXiv:2103.11436}, 2021.

\bibitem[Du et~al.(2020)Du, Yang, Zou, and Hu]{fairDL}
Du, M., Yang, F., Zou, N., and Hu, X.
\newblock Fairness in deep learning: A computational perspective.
\newblock \penalty0 (arXiv:1908.08843), Mar 2020.
\newblock URL \url{http://arxiv.org/abs/1908.08843}.

\bibitem[Goodfellow et~al.(2013)Goodfellow, Erhan, Carrier, Courville, Mirza,
  Hamner, Cukierski, Tang, Thaler, Lee, et~al.]{FER2013}
Goodfellow, I.~J., Erhan, D., Carrier, P.~L., Courville, A., Mirza, M., Hamner,
  B., Cukierski, W., Tang, Y., Thaler, D., Lee, D.-H., et~al.
\newblock Challenges in representation learning: A report on three machine
  learning contests.
\newblock In \emph{Neural Information Processing: 20th International
  Conference, ICONIP 2013, Daegu, Korea, November 3-7, 2013. Proceedings, Part
  III 20}, pp.\  117--124. Springer, 2013.

\bibitem[Hamm et~al.(2011)Hamm, Kohler, Gur, and Verma]{AU}
Hamm, J., Kohler, C.~G., Gur, R.~C., and Verma, R.
\newblock Automated facial action coding system for dynamic analysis of facial
  expressions in neuropsychiatric disorders.
\newblock \emph{Journal of Neuroscience Methods}, 200\penalty0 (2):\penalty0
  237--256, 2011.
\newblock ISSN 0165-0270.
\newblock \doi{https://doi.org/10.1016/j.jneumeth.2011.06.023}.

\bibitem[Hasani \& Mahoor(2017)Hasani and Mahoor]{Hasani_2017}
Hasani, B. and Mahoor, M.~H.
\newblock Facial expression recognition using enhanced deep 3d convolutional
  neural networks.
\newblock In \emph{2017 {IEEE} Conference on Computer Vision and Pattern
  Recognition Workshops ({CVPRW})}. {IEEE}, Jul 2017.
\newblock \doi{10.1109/cvprw.2017.282}.

\bibitem[Karkkainen \& Joo(2021)Karkkainen and Joo]{karkkainenfairface}
Karkkainen, K. and Joo, J.
\newblock Fairface: Face attribute dataset for balanced race, gender, and age
  for bias measurement and mitigation.
\newblock In \emph{Proceedings of the IEEE/CVF Winter Conference on
  Applications of Computer Vision}, pp.\  1548--1558, 2021.

\bibitem[Khorrami et~al.(2015)Khorrami, Paine, and Huang]{khorrami2015deep}
Khorrami, P., Paine, T., and Huang, T.
\newblock Do deep neural networks learn facial action units when doing
  expression recognition?
\newblock In \emph{Proceedings of the IEEE international conference on computer
  vision workshops}, pp.\  19--27, 2015.

\bibitem[Ko(2018)]{Ko2018ABR}
Ko, B.
\newblock A brief review of facial emotion recognition based on visual
  information.
\newblock \emph{Sensors (Basel, Switzerland)}, 18, 2018.

\bibitem[Li \& Deng(2018)Li and Deng]{FERsurvey}
Li, S. and Deng, W.
\newblock Deep facial expression recognition: A survey.
\newblock 2018.
\newblock \doi{10.48550/ARXIV.1804.08348}.
\newblock URL \url{https://arxiv.org/abs/1804.08348}.

\bibitem[Li et~al.(2017)Li, Deng, and Du]{RAF-DB}
Li, S., Deng, W., and Du, J.
\newblock Reliable crowdsourcing and deep locality-preserving learning for
  expression recognition in the wild.
\newblock In \emph{2017 IEEE Conference on Computer Vision and Pattern
  Recognition (CVPR)}, pp.\  2584--2593, 2017.
\newblock \doi{10.1109/CVPR.2017.277}.

\bibitem[LoBue \& Thrasher(2015)LoBue and Thrasher]{CAFE_data}
LoBue, V. and Thrasher, C.
\newblock The child affective facial expression (cafe) set: validity and
  reliability from untrained adults.
\newblock \emph{Front. Psychol.}, 5, 2015.

\bibitem[Lopes et~al.(2017)Lopes, {de Aguiar}, {De Souza}, and
  Oliveira-Santos]{LOPES2017610}
Lopes, A.~T., {de Aguiar}, E., {De Souza}, A.~F., and Oliveira-Santos, T.
\newblock Facial expression recognition with convolutional neural networks:
  Coping with few data and the training sample order.
\newblock \emph{Pattern Recognition}, 61:\penalty0 610--628, 2017.
\newblock ISSN 0031-3203.
\newblock \doi{https://doi.org/10.1016/j.patcog.2016.07.026}.

\bibitem[Lucey et~al.(2010)Lucey, Cohn, Kanade, Saragih, Ambadar, and
  Matthews]{CK_data}
Lucey, P., Cohn, J.~F., Kanade, T., Saragih, J., Ambadar, Z., and Matthews, I.
\newblock The extended cohn-kanade dataset (ck+): A complete dataset for action
  unit and emotion-specified expression.
\newblock In \emph{2010 IEEE Computer Society Conference on Computer Vision and
  Pattern Recognition - Workshops}, pp.\  94--101, 2010.
\newblock \doi{10.1109/CVPRW.2010.5543262}.

\bibitem[Lyons et~al.(1998)Lyons, Akamatsu, Kamachi, and Gyoba]{JAFFE_data}
Lyons, M., Akamatsu, S., Kamachi, M., and Gyoba, J.
\newblock Coding facial expressions with gabor wavelets.
\newblock In \emph{Proceedings Third IEEE International Conference on Automatic
  Face and Gesture Recognition}, pp.\  200--205, 1998.
\newblock \doi{10.1109/AFGR.1998.670949}.

\bibitem[Majumder et~al.(2019)Majumder, Poria, Hazarika, Mihalcea, Gelbukh, and
  Cambria]{majumder2019dialoguernn}
Majumder, N., Poria, S., Hazarika, D., Mihalcea, R., Gelbukh, A., and Cambria,
  E.
\newblock Dialoguernn: An attentive rnn for emotion detection in conversations,
  2019.

\bibitem[Martínez-Plumed et~al.(2019)Martínez-Plumed, Ferri, Nieves, and
  Hernández-Orallo]{Martínez_2019}
Martínez-Plumed, F., Ferri, C., Nieves, D., and Hernández-Orallo, J.
\newblock Fairness and missing values.
\newblock \penalty0 (arXiv:1905.12728), May 2019.
\newblock URL \url{http://arxiv.org/abs/1905.12728}.

\bibitem[Mehrabi et~al.(2022{\natexlab{a}})Mehrabi, Morstatter, Saxena, Lerman,
  and Galstyan]{Mehrabi_2022}
Mehrabi, N., Morstatter, F., Saxena, N., Lerman, K., and Galstyan, A.
\newblock A survey on bias and fairness in machine learning.
\newblock \penalty0 (arXiv:1908.09635), Jan 2022{\natexlab{a}}.
\newblock URL \url{http://arxiv.org/abs/1908.09635}.

\bibitem[Mehrabi et~al.(2022{\natexlab{b}})Mehrabi, Morstatter, Saxena, Lerman,
  and Galstyan]{biasDLSurvey}
Mehrabi, N., Morstatter, F., Saxena, N., Lerman, K., and Galstyan, A.
\newblock A survey on bias and fairness in machine learning.
\newblock \penalty0 (arXiv:1908.09635), Jan 2022{\natexlab{b}}.
\newblock URL \url{http://arxiv.org/abs/1908.09635}.

\bibitem[Mollahosseini et~al.(2017)Mollahosseini, Hasani, and
  Mahoor]{affectnet_data}
Mollahosseini, A., Hasani, B., and Mahoor, M.~H.
\newblock Affectnet: A database for facial expression, valence, and arousal
  computing in the wild.
\newblock \emph{IEEE Transactions on Affective Computing}, 2017.

\bibitem[Oneto \& Chiappa(2020)Oneto and Chiappa]{Oneto_Chiappa_2020}
Oneto, L. and Chiappa, S.
\newblock \emph{Fairness in Machine Learning}, volume 896, pp.\  155–196.
\newblock 2020.
\newblock \doi{10.1007/978-3-030-43883-8_7}.
\newblock URL \url{http://arxiv.org/abs/2012.15816}.

\bibitem[Peng et~al.(2020)Peng, Li, and Sun]{peng2020}
Peng, Z., Li, J., and Sun, Z.
\newblock Emotion recognition using generative adversarial networks.
\newblock In \emph{2020 International Conference on Computer Engineering and
  Intelligent Control (ICCEIC)}, pp.\  77--80, 2020.
\newblock \doi{10.1109/ICCEIC51584.2020.00023}.

\bibitem[Picard(1999)]{picard1999affective}
Picard, R.~W.
\newblock Affective computing for hci.
\newblock In \emph{HCI (1)}, pp.\  829--833. Citeseer, 1999.

\bibitem[Porcu et~al.(2020)Porcu, Floris, and Atzori]{porcu}
Porcu, S., Floris, A., and Atzori, L.
\newblock Evaluation of data augmentation techniques for facial expression
  recognition systems.
\newblock \emph{Electronics}, 9\penalty0 (11), 2020.
\newblock ISSN 2079-9292.
\newblock \doi{10.3390/electronics9111892}.
\newblock URL \url{https://www.mdpi.com/2079-9292/9/11/1892}.

\bibitem[Raina et~al.(2022)Raina, Monares, Xu, Fabi, Xu, Li, Sumerfield, Gan,
  and de~Sa]{raina2022exploring}
Raina, R., Monares, M., Xu, M., Fabi, S., Xu, X., Li, L., Sumerfield, W., Gan,
  J., and de~Sa, V.~R.
\newblock Exploring biases in facial expression analysis.
\newblock In \emph{NeurIPS 2022 Workshop on Synthetic Data for Empowering ML
  Research}, 2022.
\newblock URL \url{https://openreview.net/forum?id=jrddoq9NDP6}.

\bibitem[Rhue(2018)]{Rhue_2018}
Rhue, L.
\newblock Racial influence on automated perceptions of emotions.
\newblock \emph{SSRN Electronic Journal}, 2018.
\newblock ISSN 1556-5068.
\newblock \doi{10.2139/ssrn.3281765}.
\newblock URL \url{https://www.ssrn.com/abstract=3281765}.

\bibitem[Saleiro et~al.(2018)Saleiro, Kuester, Stevens, Anisfeld, Hinkson,
  London, and Ghani]{2018aequitas}
Saleiro, P., Kuester, B., Stevens, A., Anisfeld, A., Hinkson, L., London, J.,
  and Ghani, R.
\newblock Aequitas: A bias and fairness audit toolkit.
\newblock \emph{arXiv preprint arXiv:1811.05577}, 2018.

\bibitem[Sham et~al.(2023{\natexlab{a}})Sham, Aktas, Rizhinashvili, Kuklianov,
  Alisinanoglu, Ofodile, Ozcinar, and Anbarjafari]{FERbiasrace}
Sham, A.~H., Aktas, K., Rizhinashvili, D., Kuklianov, D., Alisinanoglu, F.,
  Ofodile, I., Ozcinar, C., and Anbarjafari, G.
\newblock Ethical ai in facial expression analysis: racial bias.
\newblock \emph{Signal, Image and Video Processing}, 17\penalty0 (2):\penalty0
  399–406, Mar 2023{\natexlab{a}}.
\newblock ISSN 1863-1703, 1863-1711.
\newblock \doi{10.1007/s11760-022-02246-8}.

\bibitem[Sham et~al.(2023{\natexlab{b}})Sham, Aktas, Rizhinashvili, Kuklianov,
  Alisinanoglu, Ofodile, Ozcinar, and Anbarjafari]{sham2023ethical}
Sham, A.~H., Aktas, K., Rizhinashvili, D., Kuklianov, D., Alisinanoglu, F.,
  Ofodile, I., Ozcinar, C., and Anbarjafari, G.
\newblock Ethical ai in facial expression analysis: Racial bias.
\newblock \emph{Signal, Image and Video Processing}, 17\penalty0 (2):\penalty0
  399--406, 2023{\natexlab{b}}.

\bibitem[Sohail et~al.(2022)Sohail, Ali, Rashid, Ahmad, Almotiri, AlGhamdi,
  Nagra, and Masood]{sohail}
Sohail, M., Ali, G., Rashid, J., Ahmad, I., Almotiri, S.~H., AlGhamdi, M.~A.,
  Nagra, A.~A., and Masood, K.
\newblock Racial identity-aware facial expression recognition using deep
  convolutional neural networks.
\newblock \emph{Applied Sciences}, 12\penalty0 (1), 2022.
\newblock ISSN 2076-3417.
\newblock \doi{10.3390/app12010088}.
\newblock URL \url{https://www.mdpi.com/2076-3417/12/1/88}.

\bibitem[Tao \& Tan(2005)Tao and Tan]{tao2005affective}
Tao, J. and Tan, T.
\newblock Affective computing: A review.
\newblock In \emph{Affective Computing and Intelligent Interaction: First
  International Conference, ACII 2005, Beijing, China, October 22-24, 2005.
  Proceedings 1}, pp.\  981--995. Springer, 2005.

\bibitem[Voss et~al.(2019)Voss, Schwartz, Daniels, Kline, Haber, Washington,
  Tariq, Robinson, Desai, Phillips, et~al.]{voss2019effect}
Voss, C., Schwartz, J., Daniels, J., Kline, A., Haber, N., Washington, P.,
  Tariq, Q., Robinson, T.~N., Desai, M., Phillips, J.~M., et~al.
\newblock Effect of wearable digital intervention for improving socialization
  in children with autism spectrum disorder: a randomized clinical trial.
\newblock \emph{JAMA pediatrics}, 173\penalty0 (5):\penalty0 446--454, 2019.

\bibitem[Washington et~al.(2021)Washington, Kalantarian, Kent, Husic, Kline,
  Leblanc, Hou, Mutlu, Dunlap, Penev, et~al.]{washington2021training}
Washington, P., Kalantarian, H., Kent, J., Husic, A., Kline, A., Leblanc, E.,
  Hou, C., Mutlu, C., Dunlap, K., Penev, Y., et~al.
\newblock Training affective computer vision models by crowdsourcing
  soft-target labels.
\newblock \emph{Cognitive computation}, 13:\penalty0 1363--1373, 2021.

\bibitem[Washington et~al.(2022)Washington, Kalantarian, Kent, Husic, Kline,
  Leblanc, Hou, Mutlu, Dunlap, Penev, et~al.]{washington2022improved}
Washington, P., Kalantarian, H., Kent, J., Husic, A., Kline, A., Leblanc, E.,
  Hou, C., Mutlu, O.~C., Dunlap, K., Penev, Y., et~al.
\newblock Improved digital therapy for developmental pediatrics using
  domain-specific artificial intelligence: Machine learning study.
\newblock \emph{JMIR Pediatrics and Parenting}, 5\penalty0 (2):\penalty0
  e26760, 2022.

\bibitem[Wexler et~al.(2020)Wexler, Pushkarna, Bolukbasi, Wattenberg, Viegas,
  and Wilson]{what_if}
Wexler, J., Pushkarna, M., Bolukbasi, T., Wattenberg, M., Viegas, F., and
  Wilson, J.
\newblock The what-if tool: Interactive probing of machine learning models.
\newblock \emph{IEEE Transactions on Visualization and Computer Graphics},
  26\penalty0 (01):\penalty0 56--65, Jan 2020.
\newblock ISSN 1941-0506.
\newblock \doi{10.1109/TVCG.2019.2934619}.

\bibitem[Xu et~al.(2020{\natexlab{a}})Xu, White, Kalkan, and
  Gunes]{biasesStudy}
Xu, T., White, J., Kalkan, S., and Gunes, H.
\newblock Investigating bias and fairness in facial expression recognition.
\newblock \penalty0 (arXiv:2007.10075), Aug 2020{\natexlab{a}}.
\newblock URL \url{http://arxiv.org/abs/2007.10075}.

\bibitem[Xu et~al.(2020{\natexlab{b}})Xu, White, Kalkan, and
  Gunes]{xu2020investigating}
Xu, T., White, J., Kalkan, S., and Gunes, H.
\newblock Investigating bias and fairness in facial expression recognition.
\newblock In \emph{Computer Vision--ECCV 2020 Workshops: Glasgow, UK, August
  23--28, 2020, Proceedings, Part VI 16}, pp.\  506--523. Springer,
  2020{\natexlab{b}}.

\bibitem[Zhao et~al.(2019)Zhao, Wang, Soleymani, Joshi, and
  Ji]{zhao2019affective}
Zhao, S., Wang, S., Soleymani, M., Joshi, D., and Ji, Q.
\newblock Affective computing for large-scale heterogeneous multimedia data: A
  survey.
\newblock \emph{ACM Transactions on Multimedia Computing, Communications, and
  Applications (TOMM)}, 15\penalty0 (3s):\penalty0 1--32, 2019.

\bibitem[{Zhongqing Jiang} et~al.(2023){Zhongqing Jiang}, {Guillermo Recio},
  {Wenhui Li}, {Peng Zhu}, {Jiamei He}, and {Werner
  Sommer}]{zhongqing_jiang_other-race_2023}
{Zhongqing Jiang}, {Guillermo Recio}, {Wenhui Li}, {Peng Zhu}, {Jiamei He}, and
  {Werner Sommer}.
\newblock The other-race effect in facial expression processing: {Behavioral}
  and {ERP} evidence from a balanced cross-cultural study in women.
\newblock \emph{Int J Psychophysiol}, 183:\penalty0 53--60, 2023.

\bibitem[Zhu et~al.(2017)Zhu, Li, Zhang, Rao, Xu, Huang, and Xu]{Zhu2017}
Zhu, X., Li, L., Zhang, W., Rao, T., Xu, M., Huang, Q., and Xu, D.
\newblock Dependency exploitation: A unified cnn-rnn approach for visual
  emotion recognition.
\newblock In \emph{Proceedings of the Twenty-Sixth International Joint
  Conference on Artificial Intelligence, {IJCAI-17}}, pp.\  3595--3601, 2017.
\newblock \doi{10.24963/ijcai.2017/503}.

\end{thebibliography}
\bibliographystyle{icml2023}



\end{document}